\documentclass[10pt,twocolumn,letterpaper]{article}

\usepackage[]{iccv}      
\usepackage{graphicx}
\usepackage{amsmath}
\usepackage{amssymb}
\usepackage{booktabs}
\usepackage{makecell}
\usepackage{multirow}
\usepackage{mdframed}
\usepackage{colortbl}
\usepackage{comment}
\usepackage{xr}
\usepackage{rotating}
\makeatletter
\let\@oldmaketitle\@maketitle
\renewcommand{\@maketitle}{\@oldmaketitle
     \centering
     \includegraphics[width=\linewidth]{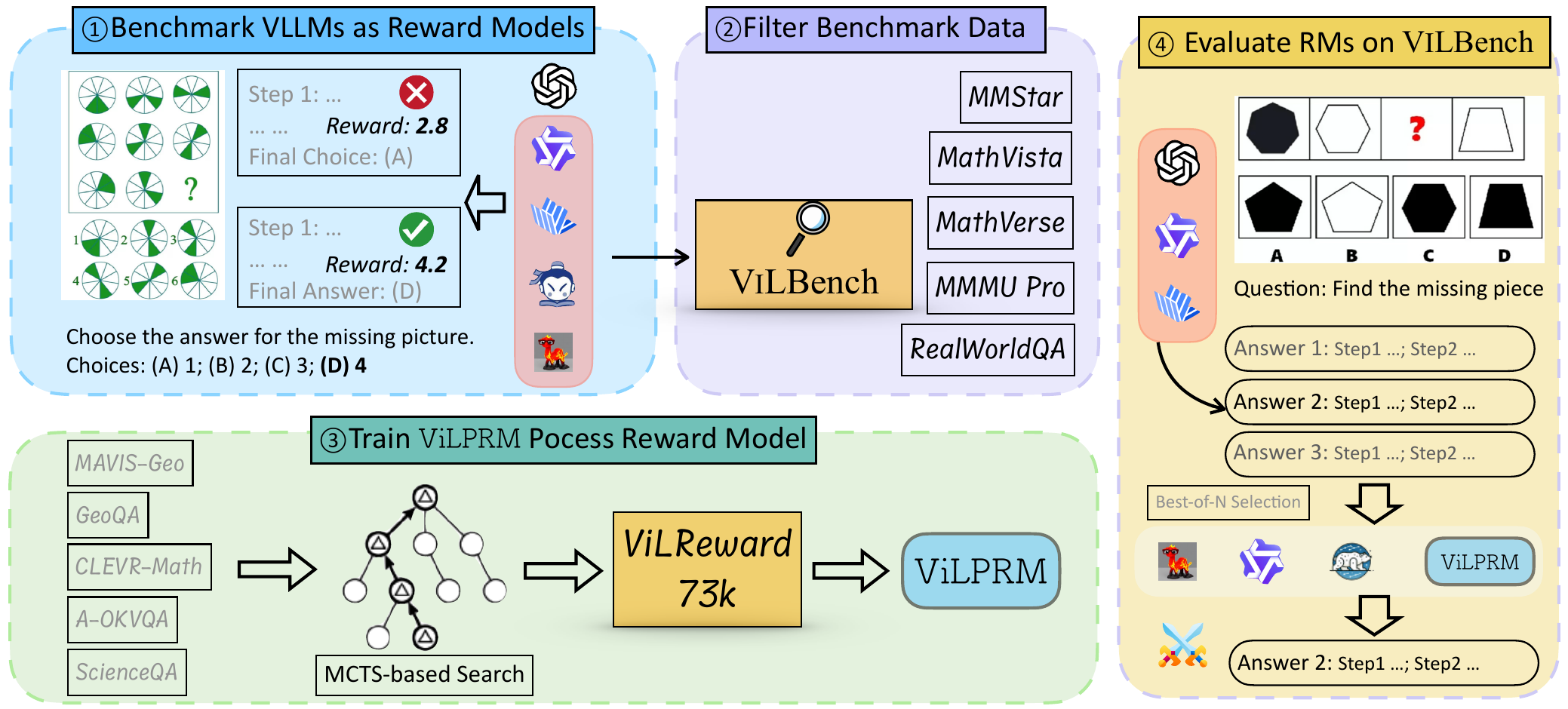}
     \vspace{-2em}
     \captionof{figure}{We present a suite of vision-language process reward modeling. We first benchmark current vision-language models as different reward models, and present \textsc{ViLBench} that requires intensive step-wise reward. Then we collect 73K+ preference reward data to train a vision-language process reward model \texttt{ViLPRM} that performs better than other baselines on \textsc{ViLBench}.}
     \label{fig:teaser}
    \bigskip}
\makeatother

\newcommand{\add}[1]{{\color{blue}{#1}}}
\newcommand{\subs}[1]{{\color{red}{#1}}}

\newcommand{\app}{\raise.17ex\hbox{$\scriptstyle\sim$}}
\makeatletter
\DeclareRobustCommand\onedot{\futurelet\@let@token\@onedot}
\def\@onedot{\ifx\@let@token.\else.\null\fi\xspace}

\def\eg{\emph{e.g}\onedot} 
\def\ie{\emph{i.e}\onedot}

\def\etal{\emph{et al}\onedot}
\makeatother

\definecolor{iccvblue}{rgb}{0.21,0.49,0.74}
\usepackage[pagebackref,breaklinks,colorlinks,allcolors=iccvblue]{hyperref}

\def\paperID{5942} 
\def\confName{ICCV}
\def\confYear{2025}

\title{\textsc{ViLBench}: A Suite for Vision-Language Process Reward Modeling}

\author{%
  Haoqin Tu$^{1}$\quad
  Weitao Feng$^{1} \thanks{Work done during Weitao's remote internship at UCSC.}$\quad 
  Hardy Chen$^{2}$\quad
  Hui Liu$^{3}$\quad 
  Xianfeng Tang$^{3}$\quad 
  Cihang Xie$^{1}$\vspace{.1em}\\
  $^{1}$UC Santa Cruz ~~ $^{2}$UT Dallas ~~ $^{3}$Amazon Research \vspace{.3em} \\
}

\begin{document}
\maketitle

\begin{abstract}
Process-supervised reward models serve as a fine-grained function that provides detailed step-wise feedback to model responses, facilitating effective selection of reasoning trajectories for complex tasks. Despite its advantages, evaluation on PRMs remains less explored, especially in the multimodal domain. 
To address this gap, this paper first benchmarks current vision large language models (VLLMs) as two types of reward models: output reward models (ORMs) and process reward models (PRMs) on multiple vision-language benchmarks, which reveal that neither ORM nor PRM consistently outperforms across all tasks, and superior VLLMs do not necessarily yield better rewarding performance.
To further advance evaluation, we introduce \textsc{ViLBench}, a vision-language benchmark designed to require intensive process reward signals. Notably, OpenAI's GPT-4o with Chain-of-Thought (CoT) achieves only 27.3\% accuracy, indicating the benchmark's challenge for current VLLMs.
Lastly, we preliminarily showcase a promising pathway towards bridging the gap between general VLLMs and reward models --- by collecting 73.6K vision-language process reward data using an enhanced tree-search algorithm, 
our 3B model is able to achieve an average improvement of 3.3\% over standard CoT and up to 2.5\% compared to its untrained counterpart on \textsc{ViLBench} by selecting OpenAI o1’s generations. We release the implementations at \url{https://ucsc-vlaa.github.io/ViLBench} with our code, model, and data.



\end{abstract}

\section{Introduction}

Reward models (RMs) play a crucial role in aligning model outputs with human preferences, benefiting Large Language Models (LLMs) in both training and inference stages~\cite{schulman2017proximal,bai2022constitutional,ouyang2022training,rafailov2023direct}. The most popular RMs include output reward models (ORMs) and process-supervised reward models (PRMs). While ORMs assess responses at the final output level~\cite{zheng2023judging,stiennon2020learning}, PRMs provide detailed, step-wise feedback, making them particularly useful for complex reasoning tasks~\cite{lightman2023let,wang2023math,zhang2025lessons}. Despite their advantages in the language domain, the application of PRMs in multimodal contexts remains underexplored, with most vision-language RMs following the ORM paradigm~\cite{xiong2024llava,lee2024prometheus,zang2025internlm,muhtar2025quality}.

To advance the study of vision-language process reward modeling, this paper presents a comprehensive suite of contributions encompassing (1) a benchmarking study of state-of-the-art VLLMs as reward models, (2) a newly curated dataset designed for fine-grained step-wise reward evaluation, and (3) an advanced vision-language PRM trained on large-scale vision-language step reward data. Our goal is to provide a deeper understanding of the effectiveness of current vision-language reward models and to pave the way for future improvements in multimodal step-wise evaluation techniques.

As our first contribution, we evaluate seven VLLMs (six open-weight and one private) following MLLM-as-a-judge~\cite{chen2024mllm,ge2023mllm} across five challenging vision-language tasks. 
This benchmarking effort systematically analyzes the models' rewarding capabilities in various domains, revealing several key insights. 
For example, we observe that neither ORM nor PRM consistently outperforms the other across all tasks, indicating that different reasoning structures benefit from different rewarding approaches~\cite{zhang2025lessons}. 
Additionally, we find that better VLLMs do not always translate to superior reward capabilities, suggesting that rewarding and generation abilities are not inherently correlated.
Our results also highlight that in specific domains such as text-dominant tasks, PRMs is able to provide a greater advantage, suggesting their strong potential in tasks requiring intricate, step-wise reasoning.

Next, we introduce \textsc{ViLBench}, a vision-language benchmark that demands step-wise reward feedback. Current vision-language benchmarks primarily focus on evaluating final outputs, which limits their ability to distinguish between improvements driven by ORMs and PRMs. To address this limitation, we curate a dataset of 600 examples that emphasize the necessity of step-wise feedback. Our filtering protocol assembles judges from six open-weight VLLMs to select examples that require fine-grained rewards beyond simple correctness assessments. Notably, advanced models like GPT-4o achieve only 27.3\% accuracy on \textsc{ViLBench}, benefiting 3.0\% more from PRM-driven step-wise rewards than from ORMs, underscoring our benchmark’s difficulty and its emphasis on fine-grained reward assessment.

Lastly, as a preliminary but promising step towards bridging the gap between general VLLMs and vision-language PRMs, we employ an enhanced multimodal Monte Carlo Tree Search (MCTS) algorithm~\cite{zhang2025rest} to generate ViLReward-73K, a dataset of 73.6K stepwise vision-language reward samples drawn from five training datasets. 
With this dataset, we train a 3B vision-language PRM that significantly improves the evaluation accuracy of step-wise rewards. Specifically, this model substantially surpasses existing PRMs, achieving an average improvement of 3.3\% over standard CoT approaches and up to 2.5\% compared to its untrained counterpart on \textsc{ViLBench}. 
We also discuss potential challenges and future directions for the development of vision-language PRMs to conclude this paper.




\begin{figure*}[ht]
    \centering
    \includegraphics[width=\linewidth]{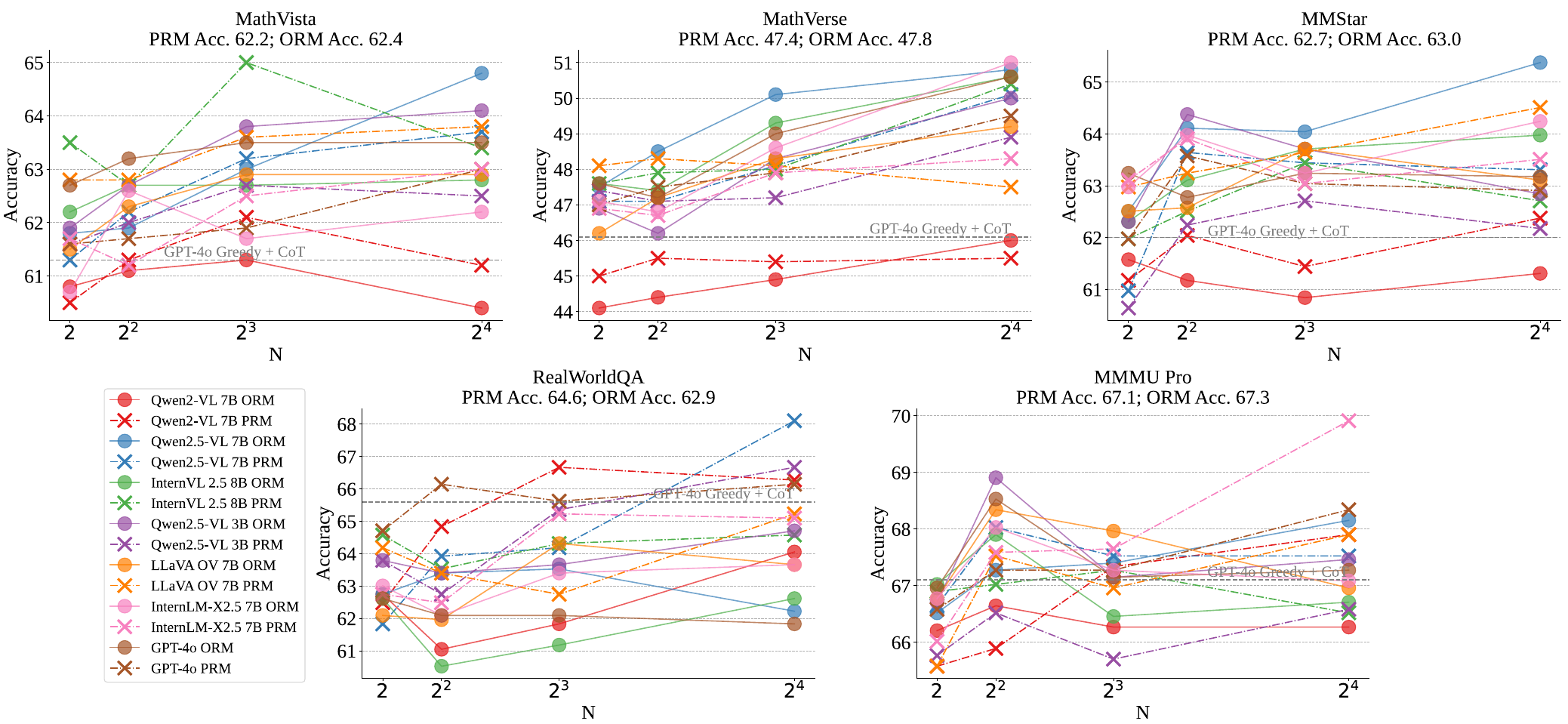}
    \vspace{-2em}
    \caption{Benchmark results of 7 different VLLMs as different reward models on 5 vision-language benchmarks. The base solution generator is GPT-4o. We report the average PRM and ORM scores in subtitles.}
    \label{fig:benchmark_overview}
\end{figure*}

\section{Part I: Benchmarking VLLMs as Reward Models}

\begin{table}[t]
    \centering
    \small
\begin{tabular}{lccc} \toprule
Model Name             & LLM & Model Size & Date     \\ \hline
InternLM-X2.5~\cite{zhang2024internlm}    &  InternLM2      & 7B         & 07/2024  \\
LLaVA-OneVisoin~\cite{li2024llava}  &  Qwen2      & 7B         & 08/2024  \\
Qwen2-VL~\cite{wang2024qwen2}         & Qwen2       & 7B         & 08/2024   \\
InternVL-2.5~\cite{chen2024expanding}     & Qwen2.5      & 8B         & 12/2024 \\
Qwen2.5-VL~\cite{bai2025qwen2}       & Qwen2.5       & 3B, 7B     & 02/2025  \\
GPT-4o~\cite{openai2024hello}           &    Unknown    & Unknown          &      05/2024    \\ \bottomrule
\end{tabular}
\vspace{-.8em}
    \caption{VLLMs used as different RMs for \textsc{ViLBench}.}
    \vspace{-1em}
    \label{tab:benchmark_model}
\end{table}

VLLMs are demonstrating increasing strength across a variety of tasks. One effective way to further enhance their performance is by evaluating their test-time scaling ability.
To assess the step-wise critique capabilities of VLLMs, we benchmark seven different models (see Table~\ref{tab:benchmark_model} for model details) following the paradigm of LLM-as-a-judge~\cite{chen2024mllm,zheng2023judging} on five widely used vision-language tasks:
MMStar~\cite{chen2024are}, MathVista~\cite{lu2024mathvista}, MathVerse~\cite{zhang2024mathverse}, MMMU Pro~\cite{yue2024mmmu}, and RealWorldQA~\cite{realworldqa}. 
To further explore their inference-time scaling potential, we adopt the Best-of-$N$ (BoN) setting, where VLLMs select the best response from a pool of $N$ candidate responses~\cite{wang2022self,lightman2023let}. 
In detail, we adopt GPT-4o (timestamp) as the base solution sampler to sample $2^4$ solutions given one question. Then we incorporate different VLLMs as the deterministic scorer to pick the best response among the candidates by assigning scores between 1 to 5 to each reasoning step. More details about prompt and model generation settings can be found in Appendix~\ref{ref:prompts}.
Through this approach, we uncover four key insights:

\begin{figure*}[ht]
    \centering
    \includegraphics[width=.9\linewidth]{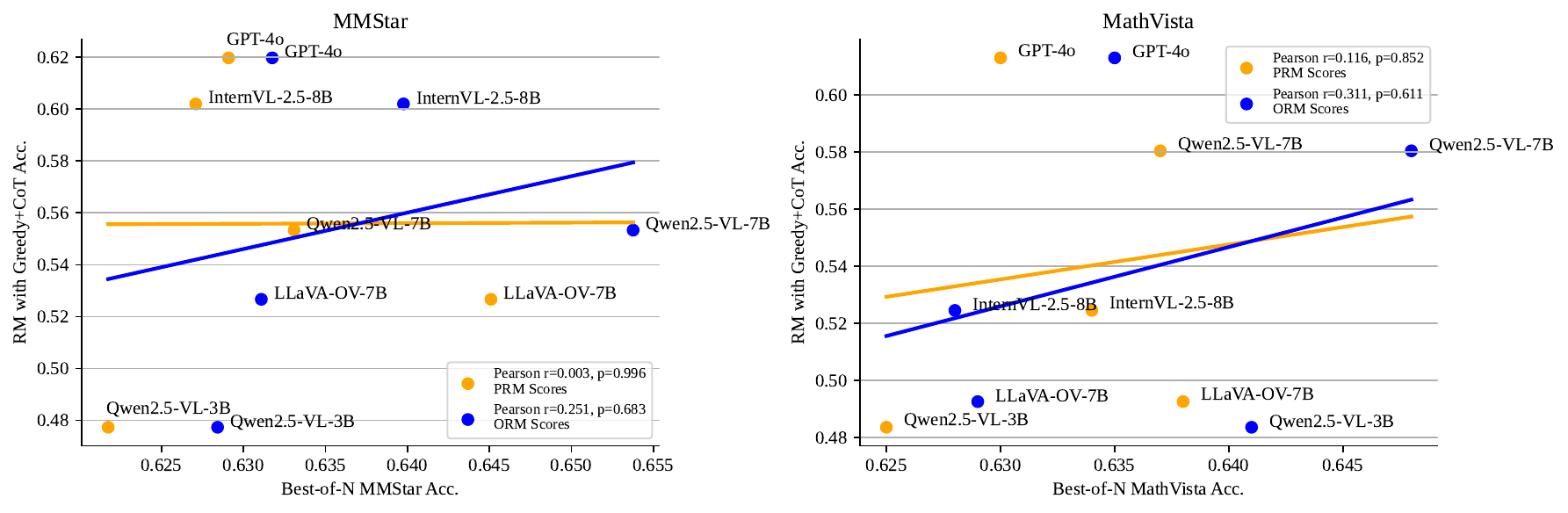}
    \vspace{-1em}
    \caption{Correlations between the model performance and its reward performance on MMStar and MathVista datasets. The rewarding performance is averaged over 4 different $N$ in the Best-of-N selection with GPT-4o as the base generator.}
    \label{fig:correlation}
\end{figure*}

\begin{figure*}[ht]
    \centering
    \includegraphics[width=.98\linewidth]{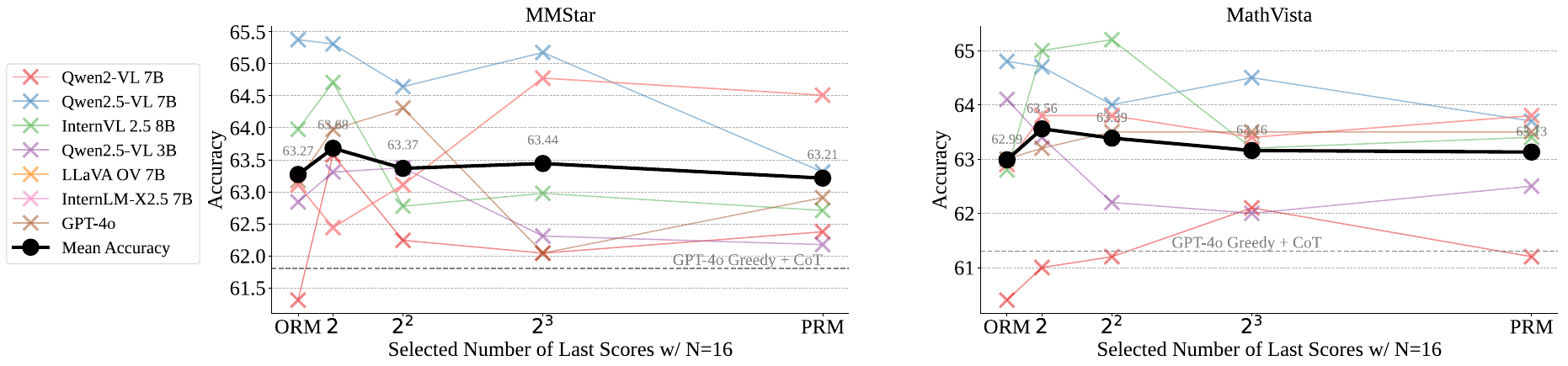}
    \caption{Model performance with the last $n$ of step rewards selected under the Best-of-N paradigm.}
    \label{fig:rms}
\end{figure*}

\begin{mdframed}[backgroundcolor=gray!15] 
\noindent\textbf{Findings 1}: Neither ORM nor PRM excels across all vision-language tasks.
\end{mdframed}
Among the five VL benchmarks in Figure~\ref{fig:benchmark_overview}, VLLMs as ORMs slightly outperform fine-grained PRMs in four cases, with an average margin of 0.3\%. 
However, on RealWorldQA, a challenging VL task involving daily life images, knowledge, and reasoning, the PRM surpasses ORM by an average of 1.7\%.
Interestingly, the four datasets where ORM performs better (MathVista, MathVerse, MMStar, and MMMU Pro) primarily feature formal reasoning and mathematical problems, whereas RealWorldQA focuses on real-world scenarios. 
This contrasts with prior findings in the language domain, where PRMs have been shown to offer better guidance than ORMs for language-only math and reasoning tasks~\cite{wang2023math,lightman2023let,wang2024openr}. 
One possible explanation is that current VLLMs are predominantly optimized on visual understanding tasks, rather than step-wise rewarding tasks.

Reward models consistently enhance performance across all five benchmarks compared to CoT greedy decoding. 
As the BoN candidate selection expands, RMs become increasingly effective in boosting performance. 
However, the impact varies across benchmarks. For example, in RealWorldQA, only four RMs at $N=2^4$ improve the base model beyond CoT. 
In contrast, for the remaining benchmarks, most RMs outperform CoT when $N>2^2$. Notably, on MathVerse, nearly all VLLMs enable GPT-4o to surpass its CoT decoding at $N\geq2$, except for LLaVA-OneVision. This observation suggests that vision-language reward models may be less effective for complex visual perception tasks than for formal reasoning challenges.

\begin{mdframed}[backgroundcolor=gray!15] 
\noindent\textbf{Findings 2}: Better vision-language models do not necessarily lead to better reward models.
\end{mdframed}
Previous research has demonstrated that stronger VLLMs tend to produce better ORMs~\cite{li2024vlrewardbench}. However, this correlation does not necessarily hold for PRMs.
To examine this, we plot the correlation between a model's greedy performance and its rewarding ability on MMStar and MathVista in Figure~\ref{fig:correlation}. 
In both tasks, ORMs exhibit higher Pearson correlation scores between general VLLM capability and rewarding ability, reinforcing the relationship between these two attributes. In contrast, under the PRM setting, the correlation is notably weak, averaging just 0.06\%. This suggests that superior VLLM performance does not directly translate to stronger rewarding capabilities, particularly in process supervision.
Notably, LLaVA-OneVision and Qwen2.5-VL rank highest as PRMs on these tasks. A surprising observation is that GPT-4o, the strongest VLLM among the tested models, underperforms as both an ORM and PRM in the deterministic scoring setting. This may be attributed to GPT-4o's tendency to overrate responses, introducing bias in certain reward tasks~\cite{song2023reward,herrera2023large}. 

Findings 1 and 2 highlight the need for the development of more robust and generalizable PRMs in the vision-language domain.

\begin{table}[t]
    \centering
    \small
    \begin{tabular}{lcc} \toprule
Method & Text-dominant & Visual-dominant  \\ \hline
Greedy & 58.9          & 51.0             \\ \hline
ORM    & 62.2 \scriptsize{\add{+3.3}}          & 49.0 \scriptsize{\subs{-2.0}}            \\
PRM    & 62.0 \scriptsize{\add{+3.1}}         & 48.9  \scriptsize{\subs{-2.1}}           \\ \bottomrule
\end{tabular}
\vspace{-.8em}
    \caption{Average performance gain across 7 RMs from text or visual dominant examples on MathVerse using ORM or PRM.}
    \label{tab:dominant}
\end{table}

\begin{mdframed}[backgroundcolor=gray!15] 
\noindent\textbf{Findings 3}: The best practice for a vision-language reward model is usually between PRM and ORM.
\end{mdframed}
Beyond PRM and ORM, alternative RMs exist that balance between selecting only the final step (ORM) and considering all step rewards (PRM) by incorporating the last $n$ step scores.
We conduct experiments using the average of the last $n$ step rewards (\ie, $n \in [1, 2, 2^2, 2^3, \text{all}]$) as the final reward signal on MMStar and MathVista. 
As shown in Figure~\ref{fig:rms}, the most effective approach consistently falls between ORM and PRM, with the optimal performance achieved by averaging the last 2 or 4 step rewards. Specifically, when using the last 2 step rewards as the final signal, the selected answer achieves the highest average accuracy, improving by 0.41\% and 0.57\% over the ORM setting on the two benchmarks.
This finding suggests an improved strategy for selecting vision-language reward signals, striking a balance between ORM and PRM for enhanced performance.

\begin{table}[t]
    \centering
    \renewcommand\arraystretch{1}
    \begin{tabular}{clccc} \toprule
                                     & Dataset Souce   & Size & Split & Ori. Size   \\ \cline{2-5}
\multirow{6}{*}{\rotatebox{90}{\textsc{ViLBench}}} & MMStar    & 150 & val  &     1,500  \\
                                     & MathVista & 150 & testmini  &    1,000    \\
                                     & MathVerse & 100 & testmini*  &    1,000    \\
                                     & MMMU Pro  & 100 & test  &    1,592     \\
                                     & RealWorldQA  & 100 & test  &    756     \\ \cline{2-5}
                                     & Sum      & 600 & test  &    5,848     \\ \bottomrule
\end{tabular}
\vspace{-.8em}
    \caption{An overview of \textsc{ViLBench}. * means that we only sample 1000 entries from testmini split of MathVerse. Ori. Size represents the original size of the dataset.}
    \label{tab:benchmark_overview}
\end{table}

\begin{mdframed}[backgroundcolor=gray!15] 
\noindent\textbf{Findings 4}: VLLMs as reward models provide more benefits on text-dominant examples.
\end{mdframed}
On MathVerse, certain examples require a stronger focus on textual reasoning (text-dominant), while others rely more on visual understanding (visual-dominant).
We report the average performance of RMs on these two subsets in Table~\ref{tab:dominant}. The results indicate that vision-language RMs provide greater benefits to VLLMs on text-dominant examples but may negatively affect performance on visual-dominant ones.
Since reward signals are integrated into the language generation process at the textual level, this finding suggests that current VLLMs exhibit stronger textual-level critique capabilities. This again, highlights the need for the development of specialized vision-language reward models to better handle visually intensive tasks.

\subsection{\textsc{ViLBench}: A Vision-Language Benchmark Requiring Intensive Reward Feedback}
As what we found previously, existing vision-language benchmarks do not require intensive feedback from RMs like vision-language PRMs. To address this, we leverage six open-weight VLLMs to filter samples where they perform well as PRMs but worse as ORMs under the BoN setting.
To be concrete, we evaluate the average performance of ORMs and PRMs using 16 response candidates, providing the RMs with a broader selection. We introduce a PRM-preference indicator based on the average PRM and ORM scores, denoted as $S_{\text{prm}}$ and $S_{\text{orm}}$, respectively. This indicator is calculated as $S = S_{\text{prm}} - S_{\text{orm}}$, allowing us to rank all samples across the five tested benchmarks according to their scores.
In Table~\ref{tab:benchmark_overview}, we illustrate how we sample varying amounts of data from each task to construct our final dataset, \textsc{ViLBench}. 
For the evaluation metric, since each question is paired with a ground truth answer, we use accuracy between predicted answers and the ground truth as the final metric, details about the answer extraction and evaluation are in Appendix~\ref{ref:evaluation}.

\section{Part II: ViLPRM: A Vision-Language Process Reward Model}

\subsection{Vision-Language Preference Data Preparation}

\begin{table}[t]
    \centering
    \resizebox{\linewidth}{!}{
    \begin{tabular}{lccc} \toprule
Dataset Source    & Class & Size & PR Size  \\ \hline
MAVIS-Geo~\cite{zhang2024mavis}  & Math & 3,093     & 25,829   \\
GeoQA170K~\cite{chen2021geoqa}  & Math & 8,063     & 31,406   \\
CLEVR-Math~\cite{lindstrom2022clevr} & Math & 957       & 1,425    \\
A-OKVQA~\cite{schwenk2022okvqa}    & General & 2,044     & 9,241    \\
ScienceQA~\cite{lu2022learn}  & General & 2,769     & 5,659    \\ \hline
Sum        & Math\&General & 16,926    & 73,560   \\ \bottomrule
\end{tabular}}
\vspace{-.8em}
    \caption{Statistics of ViLReward-73K, a vision-language process reward preference dataset. We show the initial size of the data source (Size) as well as the size of the process reward instance (PR Size).}
    \label{tab:mcts_data}
\end{table}

\paragraph{Data Selection and Filtering.}
Process preference data have been proven to be effective in training RMs in specific domains like math and logical reasoning. However, in scenarios that demand challenging visual perception understandings, the potential of PRMs remains underexplored.
In order to generalize the vision-language PRM to subjects other than just math, we consider collecting challenging VL data from general visual perception and math datasets. 
We follow three rules to filter data for the process reward model training:
\begin{enumerate}
    \item Unique image content for diverse visual features.
    \item Challenging questions that elicit model reasoning for better process scoring.
    \item Diverse source of the data for generalizing RM abilities in various domains.
\end{enumerate}

\begin{figure}[ht]
    \centering
    \includegraphics[width=1.0\linewidth]{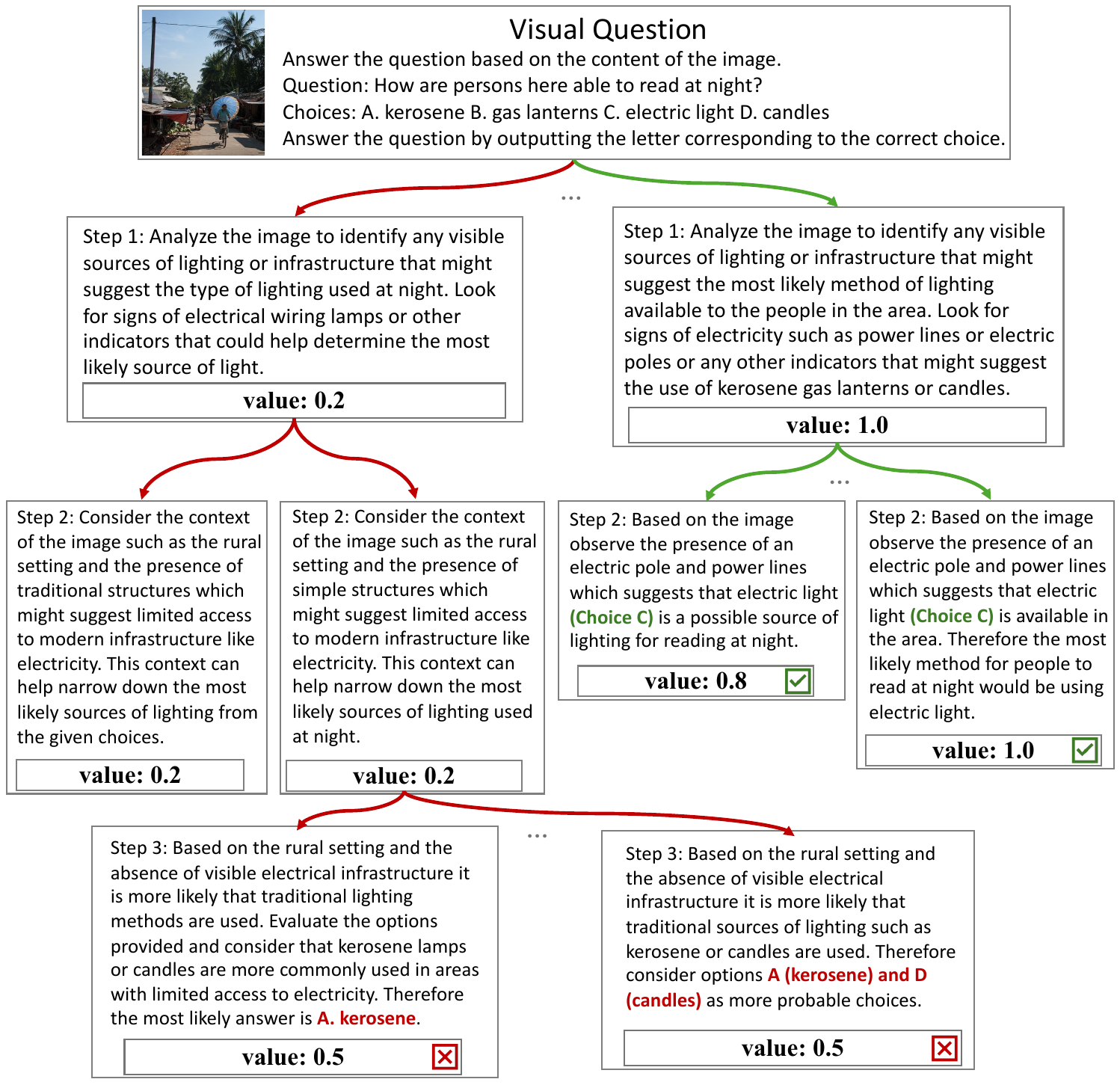}
    \vspace{-1.5em}
    \caption{An example of partial MCTS tree we constructed for ViLReward-73K. The metadata is from A-OKVQA. We mark value scores from the preference data at each node.}
    \label{fig:mcts1}
\end{figure}

In detail, we draw samples from 5 vision-language datasets consisting of 3 vision-language math data and 2 challenging visual perception tasks. 
For math domain:
\begin{itemize}
    \item MAVIS-Geometry~\cite{zhang2024mavis} is a dataset consisting of visual geometry questions that use GPT-4 to rewrite or generate geometry visual problems and solutions. There are four different difficulty levels, and we select the hardest two levels to sample 5,000 data as our metadata.
    \item GeoQA170K~\cite{gao2023g} contains over 170K geometric image-caption and question-answer pairs, building on GeoQA+~\cite{cao2022augmented} and GeoQA3K~\cite{lu2021inter}. We sample one question from each unique images from the data, resulting in 8,063 examples in total.
    \item CLEVR-Math~\cite{lindstrom2022clevr} is a synthesized VQA dataset based on CLEVR~\cite{johnson2017clevr} that includes math word problem solving. We only consider 957 questions with distinct images and need multi-hop reasoning in the dataset.
\end{itemize}
And for the visual perception domain:
\begin{itemize}
    \item A-OKVQA~\cite{schwenk2022okvqa} contains question-answering problems about natural images, we select the questions that cannot be answered directly (1,544 examples) and sample 500 data that is straightforward to answer.
    \item ScienceQA~\cite{lu2022learn} is a comprehensive dataset with 21K examples in science, the data is categorized into 12 grades based on the difficulty level. We use data that is harder than grade 7 as our metadata.
\end{itemize}
We present the detailed data volumes in Table~\ref{tab:mcts_data}.

\paragraph{MCTS Data Searching Engine.}
Instead of assigning coarse-grained binary scores (\eg, ``good'' or ``bad'') to each process~\cite{lightman2023let,wang2023math}, we adopt the reward calculation method of ReST-MCTS*~\cite{zhang2025rest} to assign fine-grained value $v$ between 0 to 1 to each reasoning step as the reward value. Unlike the previous text-based ReST-MCTS*, we have enabled visual input for the policy model, allowing the model to answer questions based on visual input. 

Following the standard MCTS tree construction, there are four major phases while tree node expansion during search: root node selection, node expansion, route simulation and value backpropagation. For the final answer evaluation, we use GPT-4o as the judge to evaluate the final output of tree search. In this part, we mainly introduce the calculation for node value and leave other details and examples in Appendix~\ref{ref:mcts_data}.
We define the quality value $v_k \in [0, 1]$  of a partial solution $ p_k = [s_1, s_2, \dots, s_k] $ to evaluate its progress toward a correct answer. $v_k$ reflects the correctness and contribution of each step \( s_i \), higher $v_k$ indicates a greater likelihood of being correct.
The reasoning distance $m_k$ is the minimum steps needed to reach the correct answer from $p_k$, estimated via simulations as it cannot be directly computed. By introducing a \emph{weighted reward} $ w_{sk} $ for step $ s_k $, incorporating $ m_k $ and the process reward $ r_{sk} $:
\begin{equation}
w_{sk} = \left(1 - \frac{v_{k-1}}{m_k + 1}\right) (1 - 2r_{sk}), \quad k = 1, 2, \dots
\end{equation}
The quality value updates iteratively:
\begin{equation}
v_k = 
\begin{cases} 
0, & k = 0, \\
\max(v_{k-1} + w_{sk}, 0), & \text{otherwise}.
\end{cases}
\end{equation}
Here, $ m_k = K - k $, where $ K $ is the total steps in solution $s$.

Since reasoning for visual problems is often simpler, we modified the prompt to require the model's output steps to be more granular, ensuring that the model does not directly output the answer at the very beginning. We present one example of the reward value tree in Figure~\ref{fig:mcts1}.



\begin{table}[t]
    \centering
    \small
    \begin{tabular}{lcccc} \toprule
Model & 3B   & 8B   & GPT-4o & o1    \\ \hline
ORM       & 30.8 & 27.8 & 28.1   & 34.9  \\
PRM       & 31.1 \scriptsize{(\add{+0.3}}) & 28.7 \scriptsize{(\add{+0.9})} & 31.1 \scriptsize{(\add{+3.0})}   & 34.9 \scriptsize{(+0.0)}  \\ \bottomrule
\end{tabular}
    \caption{The average accuracy of 3 open-weight VLLMs as reward models on the proposed benchmark. PRMs have more advantage in enhancing model performance than ORMs on our \textsc{ViLBench}.}
    \label{tab:vilbench_prove}
\end{table}

\subsection{Process Reward Model Training}
Based on the derived ViLReward-73K preference data, we train a 3B vision-language PRM, named \texttt{ViLPRM}.

\paragraph{Model Architecture.}
\texttt{ViLPRM} is built upon a 3B VLLM QwenVL-2.5~\cite{bai2025qwen2}. We follow the common practice of PRM to use the pre-trained weights of QwenVL-2.5 for most of the parts, such as the visual encoder and the MLP projector, but append a linear layer to output a scalar score after the language head~\cite{dong2024rlhf,wang2024rovrm}. We do not consider generative score modeling due to efficiency concerns.
Since the base model QwenVL-2.5 has been aligned with massive visual-language data, our reward model only requires learning to classify good or bad steps in vision-language reasoning trajectories and avoids using other pre-training data for modality alignment. 
We formalize the model input and output as: given the input question $x$ and the model reasoning response $y$, the score head $f$ transforms the logits feature of the last token into a scalar $r(x, y)$. This scalar value $r(x, y)$ serves as the predicted reward score for the inputs.

\paragraph{Training.}
Unlike previous works that only assign binary scores to the RM input, we have detailed scores for each input that can classify step responses better. We use the vanilla Mean Square Error (MSE) loss between the ground truth reward and the predicted one to update the \texttt{ViLPRM}. We add more details about training in Appendix~\ref{ref:vlprm_training}.



\section{Validating \texttt{ViLPRM} on \textsc{ViLBench}}
To validate the effectiveness of \texttt{ViLPRM}, we conduct experiments under various settings on \textsc{ViLBench}.
\subsection{Experimental Setups}
We choose the test-time scaling strategy to verify the functionality of different RMs.
On our filtered \textsc{ViLBench}, we first use 4 different VLLMs with various model size as the solution sampler, \ie, Qwen2.5-VL-3B~\cite{bai2025qwen2}, InternVL-2.5-8B~\cite{chen2024expanding}, GPT-4o~\cite{openai2024hello} and o1~\cite{openai2025o1}.
For all models except o1, we sample 16 candidate responses for BoN selection. While we sample 4 solutions for o1 due to the high cost of this sampling process.

For PRMs, we include four VLLMs --- Qwen2.5-VL-3B~\cite{bai2025qwen2}, Qwen2.5-VL-7B~\cite{bai2025qwen2}, LLaVA-OneVision-7B~\cite{li2024llava} and a recent vision-language PRM URSA-RM (URSA for short)~\cite{luo2025ursa} as baselines, where URSA is a concurrent vision-language PRM developed using a base model of 8B parameters and trained on over 1000K carefully designed preference reward data. It has shown great improvements in the multimodal math problems, but lacks the capacity to generalize to more general vision-language tasks.

\begin{figure*}[t]
    \centering
    \includegraphics[width=\linewidth]{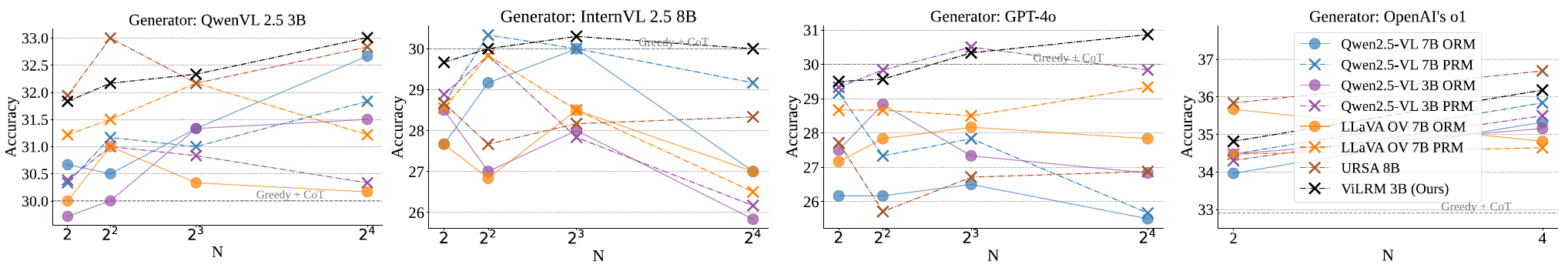}
    \vspace{-2em}
    \caption{Accuracy results of different VLLMs using various RMs under the best-of-n strategy on \textsc{ViLBench}.}
    \label{fig:vilrm_results}
\end{figure*}

\subsection{Results and Analysis}

\paragraph{\textsc{ViLBench} requires more intensive reward feedback than other VL benchmarks.}
To confirm that our filtered \textsc{ViLBench} requires more fine-grained step rewards beyond simple output rewards, we present the average accuracy across three different VLLMs (QwenVL2.5 3B, LLaVA-OneVision 7B, and QwenVL2.5 7B) used as ORMs or PRMs with four solution samplers in Table~\ref{tab:vilbench_prove}.
In most cases, PRMs enhance model performance more effectively than ORMs, yielding an average improvement of 1.4\%. However, one notable exception occurs when selecting o1's responses, where no significant difference is observed between ORM and PRM. This may be due to o1's final output steps lacking sufficient detail and accuracy, with its internal reasoning process hidden from users, making it less effective for prompt engineering~\cite{openai2025o1}.

\paragraph{\texttt{ViLPRM} performs better than other VL PRMs.}


Figure~\ref{fig:vilrm_results} presents the performance of RMs across four Best-of-N (BoN) settings with different solution samplers. As the number of response candidates increases, RMs generally enhance model performance.
Specifically trained vision-language PRMs consistently improve results over the CoT strategy for $N\geq 2^3$. 
However, VLLM-based RMs may negatively impact performance when $N$ becomes too large. For instance, among three sets of responses, three VLLMs acting as RMs exhibit varying degrees of degeneration in candidate selection from $2^3$ to $2^4$. In contrast, the two vision-language PRMs demonstrate greater consistency in identifying superior responses. This finding reinforces our previous claim that current VLLMs are not yet robust enough to serve effectively as reward models.

Table~\ref{tab:vilbench_avg} reports the average performance of different RMs. Compared to the larger PRM URSA, which is trained on over ten times more data, our \texttt{ViLPRM} achieves a superior average performance by 0.9\% and outperforms its untrained 3B VLLM counterpart by 1.3\%. Additionally, RMs consistently enhance the performance of o1, a model known for leveraging an internal thinking process as an efficient test-time scaling technique. This further underscores the importance of developing reliable reward models, even for models with built-in reasoning capabilities.

\begin{table}[t]
    \centering
    \small
    \begin{tabular}{lccccc} \toprule
Model       & 2    & 4    & 8    & 16  & Avg.  \\ \hline
QwenVL 2.5 3B~\cite{bai2025qwen2}   & 30.7 & 31.5 & 29.7 & 28.8 &  30.2 \\
LLaVA OV 7B~\cite{li2024llava} & 30.7 & 31.2 & 29.7 & 29.0 &  30.2 \\
QwenVL 2.5 7B~\cite{bai2025qwen2}   & 29.0 & 30.8 & 29.0 & 26.7 &  28.9 \\
URSA~\cite{luo2025ursa}        & 31.0 & 30.8 & 30.0 & 30.3 &  30.6 \\
\texttt{ViLPRM} (Ours)     & \textbf{31.5} & \textbf{32.0} & \textbf{31.0} & \textbf{31.3} & \textbf{31.5}  \\ \bottomrule
\end{tabular}
\vspace{-.8em}
    \caption{The average accuracy over four solution generators using different PRMs under the different BoN setups. }
    \label{tab:vilbench_avg}
\end{table}

\begin{figure}[t]
    \centering
    \includegraphics[width=.9\linewidth]{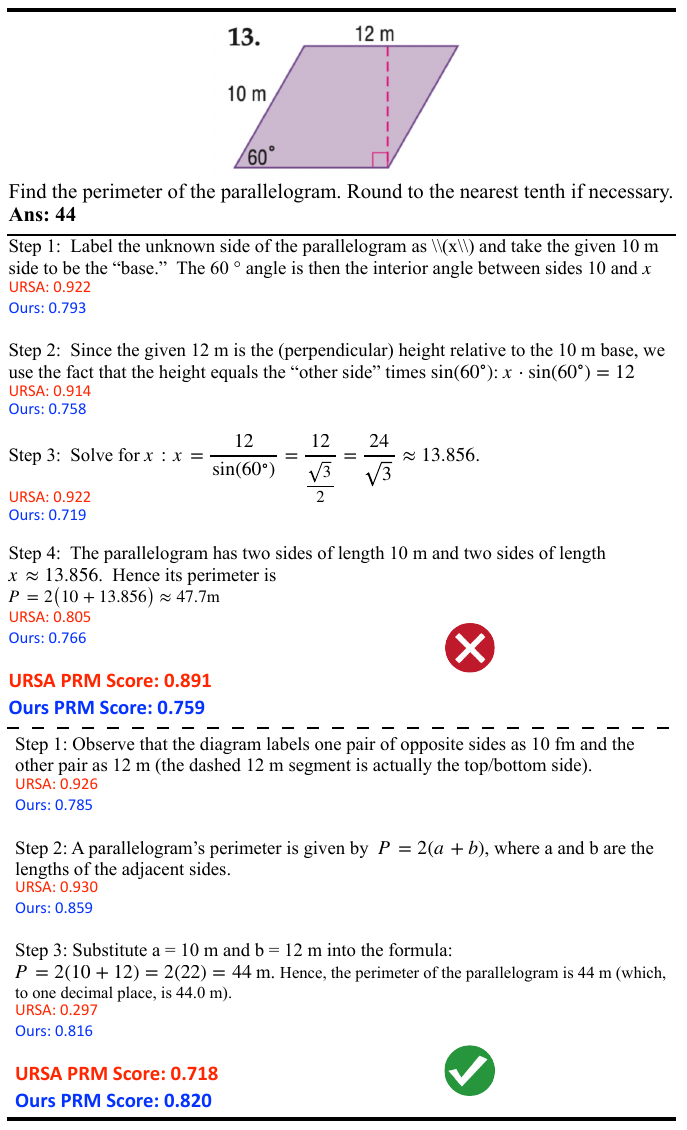}
    \caption{An example of process scores provided by URSA~\cite{luo2025ursa} and our \texttt{ViLPRM}. We mark different scores with different colors.}
    \label{fig:prm_example1}
\end{figure}
\paragraph{Examples for the Reward Task.}
In Figure~\ref{fig:prm_example1}, we present an example of using our \texttt{ViLPRM} and URSA for selecting the best response from OpenAI o1's responses. 
In the example, URSA prefers more steps in the reasoning and even with wrong trajectories, while our \texttt{ViLPRM} can choose accurate solutions. This is likely because URSA was trained on a massive amount of math reasoning data and may develop the preference for complex rather than accurate reasoning steps~\cite{liu2025can}. 
We also present another example about medical reasoning in the Appendix~\ref{ref:prm_exmaple}, which verifies that the proposed vision-language PRM has the capacity to also perform well beyond just math or reasoning tasks.

\section{Related Works}
\paragraph{Reward Benchmark.}
There are plenty of works put their emphasis in the text-only reward benchmarks~\cite{lambert2024rewardbench,liu2024rm,zhou2024rmb}, and some of them are specifically designed for PRMs~\cite{song2025prmbench,zhang2025lessons}. 
When shifting to vision-language domain, traditional VLLM evaluation mainly focuses on the general abilities of the model, including multiple aspects like knowledge, reasoning, fairness, and safety~\cite{lee2024vhelm,tu2023many,yue2024mmmu,lu2024mathvista,zhang2024mathverse,chen2024are}.  VL-RewardBench~\cite{li2024vlrewardbench} is the first work that sources reinforcement learning preference data and rewrites knowledge-intensive vision-language samples to form a diverse benchmark.
To generalize beyond just general VQA tasks, multimodal RewardBench~\cite{yasunaga2025multimodal} provides a comprehensive and high-quality benchmark for VLLM reward models, covering six key areas---general correctness, general preference, knowledge, reasoning, safety, and VQA---annotated by domain experts.
Our proposed \textsc{ViLBench} fills the gap of benchmarking PRMs in vision-language domain.

\paragraph{Reward Modeling.}

Reward models are important for guiding AI models at both training and inference stages. There are typically three different forms of RMs: (1) discriminative RM treats the rewarding task as token classification. It usually leverages a linear head to fit the reward score via a regression loss~\cite{stiennon2020learning,ouyang2022training}. (2) LLM-as-a-judge leverages the generative ability of language models to output feedback in the form of text, often a critique or explanation of why a certain output is good or bad~\cite{xiong2024llava,lee2024prometheus,zheng2023judging}. (3) Implicit RMs that are models optimized using DPO that the predicted log probabilities are interpreted as implicit reward signal~\cite{lambert2024t,ivison2023camels,zhou2024aligning}. In our work, VLLM-as-a-judge belongs to the second category, while our proposed \texttt{ViLPRM} is a discriminative RM.

Recent studies have also turned their interests from pure language to vision-language and other modalities. The initial development of vision-language RMs always falls into the specific using scenarios or inferior performance~\cite{wang2024rovrm,xiong2024llava}. Most recent works explore both the directions of VLLM-as-a-judge~\cite{xiong2024llava,ge2023mllm,chen2024mllm} and deterministic rewarding~\cite{luo2025ursa,zang2025internlm}. However, most of the current RMs are output reward models.
Liu \etal~\cite{liu2024diving} developed the first vision-language PRM in math domain, and leverage such technique for model reinforcement learning. URSA~\cite{luo2025ursa} as a concurrent work collects over 1 million vision-language reward data for training a 8B PRM that enhances math VLLMs greatly. A concurrent work VisualPRM~\cite{wang2025visualprm} collects 400K reward data for PRM training in vision language and shows great potential in both vision language and language-only reward tasks. Our \texttt{ViLPRM} breaks the tradition of employing solely on the math domain and generalizes to more general visual perception cases.

\section{Discussions and Conclusion}

\paragraph{Vision-Language PRM is Bounded by Clear Step Segmentation.}
How to best split the reasoning step for PRMs has always been a problem~\cite{liu2025can,guo2025deepseek,cui2025process}. In structured tasks like math problems, PRMs provide fine-grained feedback, improving step-by-step reasoning. However, when the segmentation of steps is unclear or reasoning is unnecessary, 
PRMs may harm the performance. 
For instance, text-heavy tasks saw a 3\% accuracy boost with PRMs, while visual-dominant tasks suffered a 2\% drop, likely due to PRMs overemphasizing irrelevant steps.

PRMs also struggle when all steps are treated equally. Previous works have proposed to use single step to represent all step rewards~\cite{wang2024openr,liu2025can}. We found that rewarding only the last few critical steps improved accuracy more than using all steps, striking a balance between PRMs and ORMs. A major challenge is identifying which steps truly matter. 
Future improvements should focus on adaptive step evaluation, where PRMs automatically adjust reward weight based on step importance. Better segmentation strategies, such as enforcing clearer step structures during training or integrating step selection mechanisms can help PRMs generalize better across tasks.

\paragraph{Improved Training Paradigm is Required for Multimodal RMs.}
Current training approaches for multimodal reward models fail to generalize across diverse tasks. Many RMs, including PRMs, are task-sensitive~\cite{zhang2025lessons,liu2025can}, meaning they work well on specific domains but struggle elsewhere. 
For example, PRMs trained on math tasks such as URSA perform poorly on vision-heavy reasoning, suggesting that current methods do not equip RMs with broad evaluation skills. Besides, our results show that even advanced VLLMs like GPT-4o do not automatically become good reward models, often overrating responses.

To improve vision-language PRMs, training must diversify data sources, integrating both textual and visual-heavy reasoning tasks. Instead of relying solely on step-wise learning, future RMs should also consider incorporating adaptive reward mechanisms, adjusting considered step scores based on task complexity. 
Additionally, evaluation benchmarks for reward models should also go beyond accuracy, assessing consistency, bias, and generalization~\cite{yasunaga2025multimodal}. 

\paragraph{Conclusion.}

We introduce \textsc{ViLBench}, a benchmark for vision-language process reward modeling (PRM), and evaluate seven VLLMs as reward models. Our findings show that PRMs enhance stepwise reasoning in structured tasks but struggle in visual-dominant scenarios, emphasizing the need for adaptive step evaluation.
To address this, we develop ViLReward-73K, a dataset of 73.6K step-wise rewards, enabling the ViLPRM to surpass other PRMs 3.3\% in accuracy. However, task sensitivity remains a challenge, highlighting the need for better reasoning step segmentation, adaptive rewards, and more diverse training data. Future work should focus on refining evaluation frameworks and improving generalization for more robust multimodal reward models.

\subsection*{Acknowledge} We thank the Microsoft Accelerate Foundation Models Research Program for supporting our computing needs.
{
    \small
    \bibliographystyle{ieeenat_fullname}
    \bibliography{main}
}

\newpage
\quad 
\newpage
\appendix
\section{Experimental Settings for VLLM-as-a-Judge} \label{ref:prompts}
In this section, we demonstrate the details of how we benchmark VLLMs as reward models on existing vision-language (VL) benchmarks. Following (V)LLM-as-a-judge paradigm, we input the pre-defined scoring rule, the question, as well as the solution steps for VLLMs to judge the score. We show our prompt below:

\begin{mdframed}[backgroundcolor=pink!15] 
You are a highly capable multimodal AI assistant tasked with evaluating the quality of intermediate reasoning steps provided for visual questions. The input answer may represent an incomplete step in the larger reasoning process. Assign a score from 1 to 5 based on how well the step contributes to addressing the question.

    \noindent Question: {question}

    \noindent Answer: {answer}

    \noindent Your score should reflect the overall quality of the answer, focusing on its relevance, coherence, accuracy, and clarity
    \noindent Scoring Scale (1-5):

    \noindent 5 (Excellent): The reasoning step is highly relevant, accurate, detailed, and exceptionally clear, making a strong contribution to addressing the question.
    
    \noindent 4 (Good): The reasoning step is relevant, mostly accurate, and clear, with logical progression and only minor flaws.
    
    \noindent 3 (Fair): The reasoning step is somewhat relevant and partially accurate, demonstrating basic logic but lacking detail, clarity, or precision.
    
    \noindent 2 (Poor): The reasoning step is partially relevant but contains major errors, lacks coherence, or is difficult to understand.
    
    \noindent 1 (Very Poor): The reasoning step is irrelevant or nonsensical, showing no meaningful connection to the question or image.

    \noindent After your evaluation, please:
    \noindent 1. Assign one overall score from 1 to 5 based on the descriptions above.
    \noindent 2. Explain your reasoning in detail, highlighting specific strengths and weaknesses of the answer.

    \noindent Example Response:
    \noindent Reasoning: [Explanation of the evaluation].
    \noindent Overall Score: [1-5]
\end{mdframed}

\begin{figure*}[t]
    \centering
    \includegraphics[width=.9\linewidth]{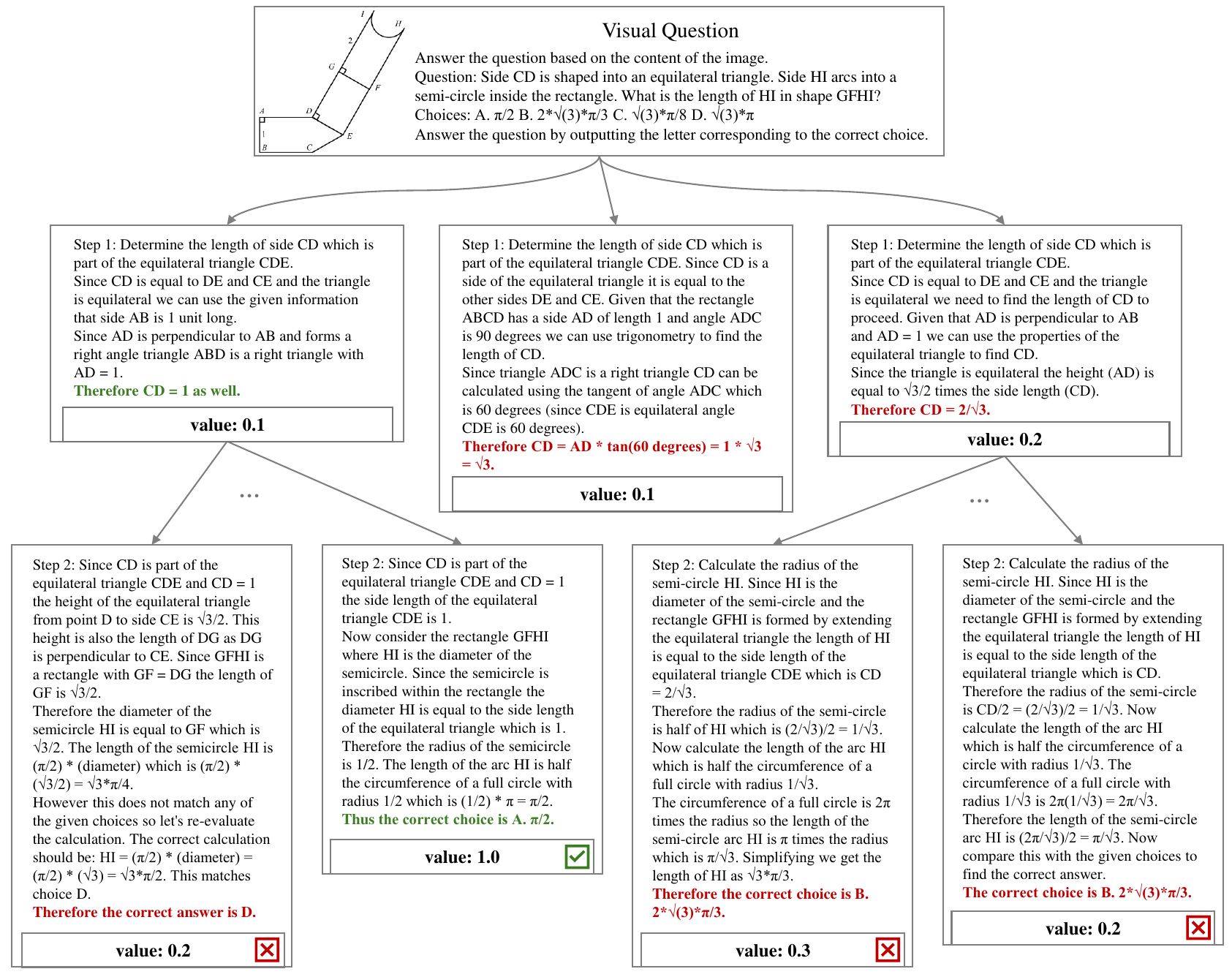}
    \caption{MCTS tree we have constructed for geometry problem datasets (\eg, MAVIS-Geometry). One path in the tree yields a correct result, while the remaining paths result in incorrect answers. It is worth noting that we use ellipses to omit some nodes in the original MCTS tree for better presentation.}
    \label{fig:MAVIS-Geometry MCTS}
\end{figure*}

\section{Data Collection Details for ViLReward-73K}\label{ref:mcts_data}
\subsection{Data Selection}
We detail the five datasets that we leverage for ViLReward-73K.
\begin{itemize}
    \item MAVIS-Geometry~\cite{zhang2024mavis} is a mathematical geometry problem dataset that includes 4 different difficulty levels, marked as depth0, depth1, depth2, and depth3. We found that depth0 and depth1 are relatively simple, while depth3, compared to depth2, mostly just increases the number of composite bodies without significantly increasing the difficulty. We chose depth2 as the source for our synthetic data, selecting 5000 examples from it as our question data. MAVIS-Geometry categorizes question types into three classes: text-dominant questions, text-lite questions, and vision-dominant questions. We use vision-dominant questions to enhance the model's visual capabilities. We demonstrate the MCTS tree constructed on MAVIS-Geometry in~\cref{fig:MAVIS-Geometry MCTS}.
    \item A-OKVQA~\cite{schwenk2022okvqa} mainly contains question-answering problems about natural images, while the majority of these problems are relatively straightforward, requiring only basic visual recognition and common knowledge. In our study, we focused specifically on the more challenging questions within this dataset. We kept 9\% of the more difficult ones which were labeled as ``difficult direct answer". To generalize to the general visual perception domain, we also sample 500 examples as metadata from the questions that can be directly answered.
    \item GeoQA170K~\cite{gao2023g} contains over 170K geometric image-caption and question-answer pairs, building on GeoQA+~\cite{cao2022augmented} and GeoQA3K~\cite{lu2021inter}. We sample one question from each unique images from the data, resulting in 8,063 examples in total.
    \item CLEVR-Math~\cite{lindstrom2022clevr} is a synthesized VQA dataset based on CLEVR~\cite{johnson2017clevr} that includes math word problem solving. They incorporate addition/subtraction types of math problems. We only consider 957 questions with distinct images and need multi-hop reasoning in the dataset.
    \item ScienceQA~\cite{lu2022learn} is a comprehensive datasets with 21K examples in science, the data is categorized into 12 grades based on the difficulty level. We use data that is harder than grade 7 as our metadata.
\end{itemize}

\subsection{MCTS Searching Details}
Our construction of the search tree is primarily based on the Monte Carlo Tree Search (MCTS). The process of building this search tree follows several key steps:

\paragraph{Search Tree Initialization.}
The search process begins with a root node that represents the initial state of the problem. This root node serves as the foundation for the entire search tree, with its parent node set to \texttt{None}.

\paragraph{Node Expansion During Search.}
During each iteration of the MCTS process, the tree undergoes expansion through four essential phases:

\begin{itemize}

\item \textbf{Selection Phase.}
Starting from the root node, a path is selected based on a specific strategy (we use the Upper Confidence Bound algorithm) until a node that has not been fully expanded is identified.

\item \textbf{Expansion Phase.}
For a node that has not been fully expanded, potential child nodes are generated. Each child node represents a possible subsequent state and is incorporated into the search tree structure with appropriate depth and parent-child relationships.

\item \textbf{Simulation Phase.}
From each newly expanded node, a simulation is conducted using a predetermined strategy (often a random approach) until a terminal state is reached.

\item \textbf{Backpropagation Phase.}
The results obtained from the simulation are propagated backward through all nodes along the path from the root to the expanded node. This process updates key node statistics including the visit count and value estimations.
\end{itemize}

\paragraph{Termination Criteria.}
The construction of the search tree continues until specific termination conditions are met. In our implementation, we set the iteration limit to 10, meaning the search process concludes after completing 10 iterations. Once this criterion is satisfied, the search process terminates, and the final tree structure is established.

\paragraph{Answer Evaluation.}
We follow the LLM-as-a-judge approach, using LLM to score the final answers and then propagate these scores from leaf nodes to previous nodes. We first prompt the LLM to extract the final answer from the model's output, then input the question, the model-generated answer, and the correct answer to the LLM to determine whether the final answer is correct.

\section{\textsc{ViLBench} Evaluation Details}\label{ref:evaluation}
We employ the accuracy between predicted answers and the ground truth as the metric for our \textsc{ViLBench}. To avoid inaccurate extraction of the answer, we follow previous works~\cite{lu2024mathvista,zhang2024mathverse} to employ GPT-based extraction. In detail, we prompt GPT-3.5-turbo to compare the prediction with the ground truth, the input instruction shows below:
\begin{mdframed}[backgroundcolor=pink!15] 
Given the following:

\noindent\#\#\# Generated Answer: {model predicted answer}

\noindent\#\#\# Ground Truth Answer: {ground truth answer}

\noindent Please compare the final answer in the generated response to the ground truth answer. Ignore any reasoning or intermediate steps and focus only on whether the final letter answer in the generated response matches the ground truth.

\noindent Output True if the final answer aligns with the ground truth answer; otherwise, output False.
\end{mdframed}

\section{Vision-Language PRM Training Details} \label{ref:vlprm_training}
We employ the value head architecture for PRM training. In detail, we train the model on our ViLReward-73K for 2 epochs with a constant learning rate of $2e^{-5}$. We randomly sample 300 instances as the validation set during training and save the model checkpoint with the lowest validation loss. 


\section{Scoring Examples from RMs} \label{ref:prm_exmaple}
\begin{figure}[t]
    \centering
    \includegraphics[width=.9\linewidth]{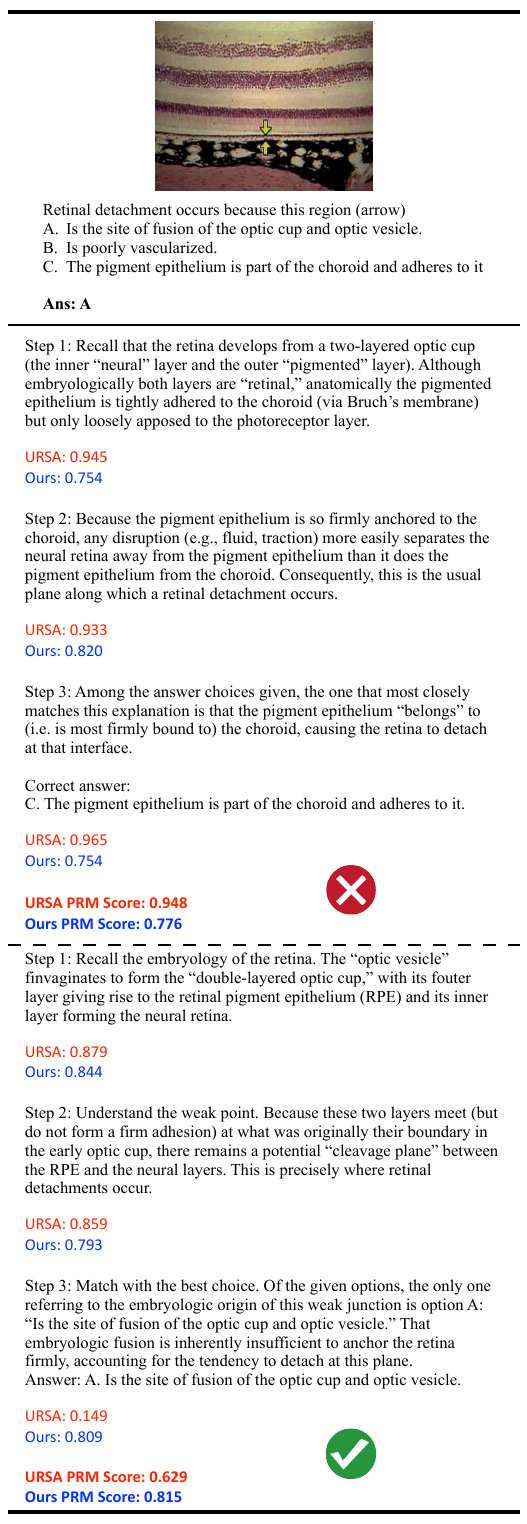}
    \caption{An example from o1's generation of a medical reasoning example from our \textsc{ViLBench}.}
    \label{fig:prm_example2}
\end{figure}

In Figure~\ref{fig:prm_example2}, we present another example in the domain of medical reasoning task. As the PRM URSA~\cite{luo2025ursa} was not trained on the domain of general knowledge, it gives biased judges in this case. In the meanwhile, our \texttt{ViLPRM} is capable of providing more accurate and consistent step rewards in this domain.
\end{document}